\RequirePackage{fix-cm}
\documentclass[twocolumn]{svjour3}          
\smartqed  
\usepackage{xcolor}
\usepackage{tabularx}
\usepackage{multirow}
\usepackage{graphicx}
\usepackage{makecell}
\usepackage{amsmath}
\usepackage{color}
\usepackage{float}
\usepackage{rotating}
\usepackage{placeins}
\usepackage{amsfonts}
\usepackage{bm}

\newcommand{\cmark}{\ding{51}}%
\newcommand{\xmark}{\ding{55}}%

\def\eg{\textit{e.g.}}
\def\ie{\textit{i.e.}}

\usepackage[title]{appendix}

\usepackage[round,authoryear,comma,sort&compress,numbers]{natbib} 
\usepackage[colorlinks=true,linkcolor=blue,citecolor=black,urlcolor=blue]{hyperref}

\usepackage{diagbox}
\usepackage{cancel} 
\usepackage{mathrsfs} 
\usepackage{eqnarray} 

\usepackage{tikz}
\usepackage{ctable} 

\usepackage{algorithm}
\usepackage{algpseudocode}%
\usepackage{listings}%
\usepackage{bbding}
\usepackage{bbm}
\usepackage[caption=false, font=footnotesize]{subfig}
\usepackage{fontawesome}
\usepackage{array}

\usepackage{colortbl}  
\usepackage{arydshln}
\usepackage{longtable}
\usepackage{wrapfig}
\usepackage{bbold}
\usepackage{caption}
\usepackage{pifont}

\usepackage{xcolor}
\usepackage{hyperref}

\hypersetup{hidelinks,
	colorlinks=true,
	allcolors=black,
	pdfstartview=Fit,
	breaklinks=true}

\definecolor{lime}{HTML}{A6CE39}
\DeclareRobustCommand{\orcidicon}{
\begin{tikzpicture}
\draw[lime, fill=lime] (0,0)
circle[radius=0.13]
node[white]{{\fontfamily{qag}\selectfont \tiny \.{I}D}};
\end{tikzpicture}
\hspace{-2mm}
}
\foreach \x in {A, ..., Z}{%
\expandafter\xdef\csname orcid\x\endcsname{\noexpand\href{https://orcid.org/\csname orcidauthor\x\endcsname}{\noexpand\orcidicon}}
}

\newcommand{\PreserveBackslash}[1]{\let\temp=\\#1\let\\=\temp}
\newcolumntype{C}[1]{>{\PreserveBackslash\centering}p{#1}}
\newcolumntype{R}[1]{>{\PreserveBackslash\raggedleft}p{#1}}
\newcolumntype{L}[1]{>{\PreserveBackslash\raggedright}p{#1}}

\def\eg{\textit{e.g.}}
\def\ie{\textit{i.e.}}

\definecolor{battleshipgrey}{rgb}{0.52, 0.52, 0.51}
\definecolor{capri}{rgb}{0.0, 0.75, 1.0}
\definecolor{mediumspringgreen}{rgb}{0.0, 0.98, 0.6}
\definecolor{Gray}{rgb}{0.7,0.7,0.7}

\journalname{%
  \parbox[t]{8cm}{%
    International Journal of Computer Vision\\
    https://doi.org/10.1007/s11263-025-02417-3%
  }%
}
\begin{document}

\title{LaneCorrect: Self-supervised Lane Detection
}


\author{
	Ming Nie$^1$ \and
	Xinyue Cai$^2$ \and
    Hang Xu$2$ \and
    Li Zhang$^1$\hspace{-2mm}\orcidA{}
}




\institute{
	Corresponding author: Li Zhang  \at
             \email{lizhangfd@fudan.edu.cn}          \\
$^1$School of Data Science, Fudan University \\
$^2$Huawei Noah’s Ark Lab 
}
\date{5th Mar 2025}

\maketitle

\begin{abstract}
Lane detection has evolved highly functional autonomous driving system to understand driving scenes even under complex environments.
In this paper, we work towards developing a generalized computer vision system able to detect lanes without using \textit{any} annotation.
We make the following contributions: 
(i) We illustrate how to perform unsupervised 3D lane segmentation by leveraging the distinctive intensity of lanes on the LiDAR point cloud frames, and then obtain the noisy lane labels in the 2D plane by projecting the 3D points;
(ii) We propose a novel self-supervised training scheme, dubbed {\em LaneCorrect}, that automatically corrects the lane label by learning geometric consistency and instance awareness from the adversarial augmentations;
(iii) With the self-supervised pre-trained model, we distill to train a student network for arbitrary target lane (\eg,~\textit{TuSimple}) detection without any human labels;
(iv) We thoroughly evaluate our self-supervised method on four major lane detection benchmarks (including \textit{TuSimple, CULane, CurveLanes} and \textit{LLAMAS}) and demonstrate excellent performance compared with existing supervised counterpart, whilst showing more effective results on alleviating the domain gap, \ie,~training on \textit{CULane} and test on \textit{TuSimple}.
\end{abstract}

\section{Introduction}

Lane detection is a fundamental task in any autonomous driving system requiring reasoning about the shape and position of marked lanes and is of great importance for path planning, steering of vehicles and line keeping for the autonomous system.
Given an image of the street scene, the objective of lane detection is to estimate the position of the lane paints with the best possible accuracy.
Existing approaches for lane detection focus on developing discriminative feature representations to classify whether each pixel represents the lane and assign it to its respective instance~\citep{SCNN,SAD}, or explicitly learn from pre-defined proposals and perform a detection task~\citep{pointlanenet,CurveLane-NAS,Tabelini2020KeepYE}, both in a supervised fashion.

Although the existing supervised methods have been shown to be effective in improving the lane detection performance on specific benchmarks over the past five years, lane detection remains an unsolved problem.
This is because that the real-world driving scenes often undergo dramatic changes due to different camera sensors, road type, changes in illumination and background clutter.
For example, a lane detector trained on the data collected in \textit{US west coast} would have problem to detect the lanes at \textit{London Piccadilly circus}.
This severely limits their scalability.
Training with more data and annotations from different types of roads even on the occluded scenario in a supervised-learning paradigm might resolve this problem, but it would inevitably increase annotation costs.

\begin{figure}[t]
    \centering
    \includegraphics[scale=0.53]{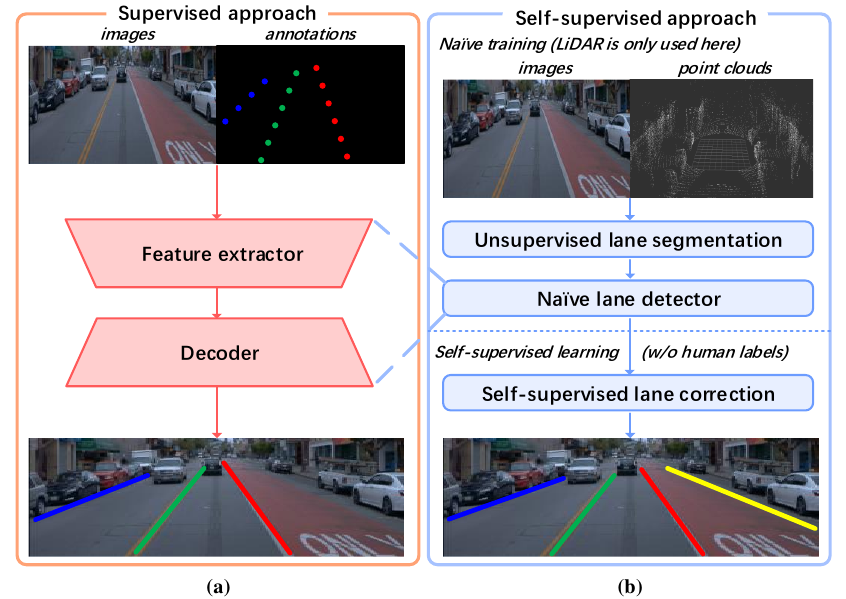}
    \caption{
    Comparison between our self-supervised lane detection method and the supervised alternative.
    (a) is the supervised approach, which relies on the human annotations as supervision.
    (b) is our \textit{LaneCorrect}, which leverages point clouds to generate noisy lane clues at first and trains a lane correction network in the self-supervised manner.
    No human annotations are introduced in our approach.}
    \label{fig:overview}
\end{figure}

In this paper, we study an self-supervised learning strategy, dubbed \textit{LaneCorrect}, to remedy above issue (see Figure~\ref{fig:overview}).
Conventional unsupervised learning methods attempt to detect lanes by using hand-crafted feature and curve fitting (\eg, hough transform~\cite{hough-transform} and B-spline fitting~\cite{b-spline}), but with little success.
Inspired by~\cite{tian2021unsupervised}, we propose to detect noisy lane paints by leveraging the large-scale LiDAR point cloud in an unsupervised manner.
The special material of road markers paint is always designed to reflect enough vehicle lights to be seen even in the poor light condition.
It also shows a distinctive reflectivity difference between bare pavement and lane paint in LiDAR point cloud.
Taking usage of this reflectivity distortion around the lane, we can extract candidate lane instances from the 3D points by using the DBSCAN~\cite{8334807} alongside RANSAC~\cite{Fischler1981RandomSC}, and then predict the lanes in the 2D image plane by projecting the 3D points.
We can use such noisy lane prediction as pseudo labels to train a \textit{naïve} lane detector.
However, the noisy labels may impact the quality of the learned lane detector.
Hence, we develop a self-supervised algorithm to leverage geometric symmetries of lanes lines and reduce the perturbations of noisy label.

In view of the geometric consistency behind the low-density separation assumption~\cite{ouali2020overview}, \ie, data points of the same cluster are likely to be of the same decision boundary, we train a self-supervised lane correction (\textit{LaneCorrect}) model with the RGB image and its noisy label cues as inputs, producing a corrected label without any knowledge of ground-truth annotation.
Since inductive geometric symmetries are inherent characteristics of lane annotation, multiple disturbances of the same lane label can be viewed as the multiple additive Gaussian noises applied to the same noise-free annotation.
Specifically, we perturb the original pseudo label with two different augmentations and enable the network to be trained and functioned as a correction with a consistency loss.
Both predictions are corrected by the network so one prediction can be utilized as another prediction's label and \textit{vice versa}.
A reconstruction loss is also added to avoid collapsing to trivial solutions~\cite{He2020MomentumCF}.
Furthermore, the lane mask pooling followed by a contrastive loss are augmented in the feature representation for instance similarity learning.
With the self-supervised pre-trained model, we distill it to train a student network on arbitrary target lane detection dataset (\eg,~\textit{TuSimple}) without touching its ground-truth label, to predict the self-supervised trained model's representation of the same image.
{\bf Note we do not rely on LiDAR at the inference phase} and the noisy label input to our {\em LaneCorrect} model comes from the prediction of the \textit{naïve} lane detector.

The \textbf{contributions} of this work are as follows: 
\textbf{(i)} 
We show that the LiDAR view for the lane instance can be utilized as pseudo labels for lane detection and propose an unsupervised 3D lane segmentation method to predict such \textit{noisy} pseudo labels;
\textbf{(ii)} A novel self-supervised training scheme for the noisy lane correction model has been formulated by learning consistency and instance awareness from different augmentations;
\textbf{(iii)} With the self-supervised pre-trained model, we distill to train a student network dedicates to predict lanes in arbitrary target datasets without using \textit{any} annotations;
\textbf{(iv)} Extensive experiments on four major lane detection benchmarks demonstrate that our model achieves on-par performance with the supervised rival (pretrained with ImageNet), whilst showing superior performance on alleviating the domain gap, \ie,~training on \textit{CULane} and test on \textit{TuSimple}.

\section{Related work}

\noindent\textbf{Unsupervised lane detection and point cloud segmentation.}
Early works on lane detection are based on unsupervised methods~\cite{dllmd} with handcrafted features.
They show poor performance and only tackle simple scenes and obvious lanes~\cite{hough-transform,b-spline}.
Existing unsupervised point cloud segmentation methods can be categorized into four gro- \
ups: edge-based~\cite{inproceedings,Bhanu1986RANGEDP,572006}, region growing~\cite{ning,interpolation,DONG2018112,Rabbani}, model fitting~\cite{osti_4746348,Fischler1981RandomSC} and clustering-based~\cite{8334807,TomoSAR,6573417,polyhedral,5651310}.
In this paper, we investigate distinctive intensity of lanes among surrounding environment under LiDAR points and propose an unsupervised 3D lane segmentation approach.
We thus obtain the initial lane predictions in 2D plane by projecting the segmented 3D points.

\noindent\textbf{Supervised lane detection}
Most existing lane detection methods are based on dense prediction approach~\cite{pan2018spatial,SAD,xu2022rclane}, which treat lane detection as a pixel-level segmentation task.
Recently there is a surge of interest in proposal-based methods~\cite{li2019line,pointlanenet,CurveLane-NAS,Tabelini2020KeepYE,zheng2022clrnet,liu2021condlanenet} to perform efficient lane detection.
In addition, there are a few row-wise detection~\cite{qin2020ultra} and parametric prediction~\cite{tabelini2021polylanenet} based methods in the literature demonstrate their superiority on lane detection problem.

\noindent\textbf{Learning with noisy annotation.}
Most existing works on training with noisy annotation employ the strategies of selecting a subset of clean labels~\cite{Nishi2021AugmentationSF,Malach2017DecouplingT,Jiang2018MentorNetLD} or leverage the output predictions of the network to correct the loss~\cite{Nishi2021AugmentationSF,Patrini2017MakingDN,Chen2019AutoCorrectDI}.
In this work, we introduce a novel self-supervised training scheme that automatically correct the lane labels by learning consistency and instance awareness from the geometry translation and rotation noise.

\noindent\textbf{Leveraging unlabeled data.}
Recently, several studies explore lane detection algorithms in semi-supervised or unsupervised forms.
\cite{garnett2020synthetic} proposes a UDA method to transfer from synthetic data to an unlabeled target domain, while \cite{lin2021semi} introduces a semi-supervised lane detection method using Hough Transform loss.
To better leverage data from native autonomous driving scenes, we propose a novel method that utilizes unlabeled data with the aid of lidar clues.
Additionally, contrastive learning has recently shown great promise~\cite{1640964,Chen2020ASF,He2020MomentumCF,Wu2018UnsupervisedFL,Asano2020SelflabellingVS,Caron2019LeveragingLU,Caron2020UnsupervisedLO,Goyal2021SelfsupervisedPO,he2022masked,gao2022mcmae} in self-supervised representation learning.
Most of the works focus on the image-level representations.
However, there has been an increasing interest in learning the instance similarity which is more effective in downstream tasks such as detection and segmentation~\cite{Gansbeke2021UnsupervisedSS,Zhang2020SelfSupervisedVR,Henaff2021EfficientVP}.
In this paper, we propose an instance similarity learning strategy in the representation level alongside consistency learning to pre-train a lane predictions model without human labels.

\section{Method}

\subsection{Overview}
Figure \ref{fig:overview} provides an overview of our \textit{LaneCorrect} strategy. 
It first takes synchronous 2D images and 3D LiDAR frames as inputs.
With the proposed unsupervised 3D lane segmentation algorithm, candidate lane instances are extracted from the 3D point clouds and then projected on the 2D image plane as the pseudo labels.
Next, considering that the pseudo label generated by the above method has a specific noise (\eg,~projection error), a self-supervised lane correction network (\textit{SLC}) is trained to reduce the noise of the pseudo labels.
Finally, we distill the \textit{SLC} model to train a student lane detector on target domain to perform lane detection task without any annotations.

\subsection{Unsupervised 3D lane segmentation} 
We first introduce our unsupervised 3D lane segmentation algorithm.
It takes 3D point clouds as input and generate 3D lane instances in LiDAR frames, as shown in Algorithm~\ref{alg:density cluster}.

\begin{algorithm}[t]
    \renewcommand{\algorithmicrequire}{\textbf{Input:}}
    \renewcommand{\algorithmicensure}{\textbf{Output:}}
    \caption{unsupervised lane instance clustering}
    \label{alg:density cluster}
    \begin{algorithmic}
    \Require laneline candidates $P_{L}=\{(p_x,p_y,p_z,p_i)\}$
    \State initialization:$L \gets \emptyset$; $i \gets 0$
    \State $L^0=\{C_{i}^{0}\}_{i=1}^{k} \gets DBSCAN(P_{L}, \epsilon_{1}, M_{1})$
    \Repeat
    \State $i \gets i + 1$
    \State calculate the center coordinates $(x_i, y_i)$ of $C^0_{i}$
    \For{$C_{j} \in L$}
        \State calculate the center coordinates $(x_j, y_j)$ of $C_{j}$
        \If{$\lvert x_i - x_j \rvert < \epsilon_{2}$ and $\lvert y_i - y_j \rvert > \epsilon_{3}$}
        \State $C_{j} \gets C_{j} \cup C^0_{i}$
        \State $L^0 \gets L^0 \setminus \{C^0_{i}\}$
        \State \textbf{break}
        \EndIf
    \EndFor
    \State $L^0 \gets L^0 \setminus \{C^0_{i}\}$
    \State $L \gets L \cup \{C^0_{i}\}$
    \Until $L^0 = \emptyset$
    \For{$C_{j} \in L$}
        \If{$count(C_{j}) < M_{2}$}
        \State $L \gets L \setminus \{C_{j}\}$
        \EndIf
    \EndFor
    \Ensure lane instance clusters $L=\{C_i\}_{i=1}^{K}$
    \end{algorithmic}
\end{algorithm}

We denote the input 3D point clouds as $P = \{(p_x,p_y,\\
p_z,p_i)\}$, where $p_x, p_y$ and $p_z$ represent the 3D point coordinates in LiDAR frame, and $p_i$ represents the intensity value of this point.
First, we use region growing strategy~\cite{ning} to segment all the input point cloud $P$, and obtain the ground point cloud $P_G \subset P$.
It is worth mentioning that lane paintings or road surface markings have distinctive intensity $p_i$ among their surrounding environment, \ie, asphalt or cement surface with low intensity for the reason that they are painted with special materials to be clearly witnessed even during the night.
Therefore, this property leads to an intensity distortion around the lane in the 3D point space.
Based on this prior, we filter the ground point clouds by setting a minimum threshold $\tau$ at the intensity to get the preliminary lane candidate points $P_L \subset P_G$.

For all lane candidate points, to generate 3D lane instance, we use DBSCAN~\cite{8334807} to cluster the segmented candidate points into $k$ clusters $\{C_i\}_i^k$, where the auto-defined $k$ represents the number of lane instances segmented by our method.

Finally, to reduce the influence of clustering noise on lane fitting, we used RANSAC~\cite{Fischler1981RandomSC} to perform curve fitting on each cluster $C_i$ to obtain 3D lane proposals $Y = \{(p_x, p_y, p_z)\}$.
These 3D lane instances $Y$ are projected to 2D frames to generate 2D lane coordinates $y = \{(p_u, p_v)\}$, where $p_u$ and $p_v$ denote the 2D pixel coordinates on 2D image plain.

Although the pseudo labels $y$ have considerable accuracy, noise inevitably exists in the annotations, \eg, projection errors in the process from 3D to 2D, as well as missed and mislabelled lanes caused by clustering errors.
Training with these noisy labels blindly damages the performance of the lane detectors seriously.

\subsection{Self-supervised lane correction network}
\begin{figure*}[t]
    \centering
    \includegraphics[scale=0.6]{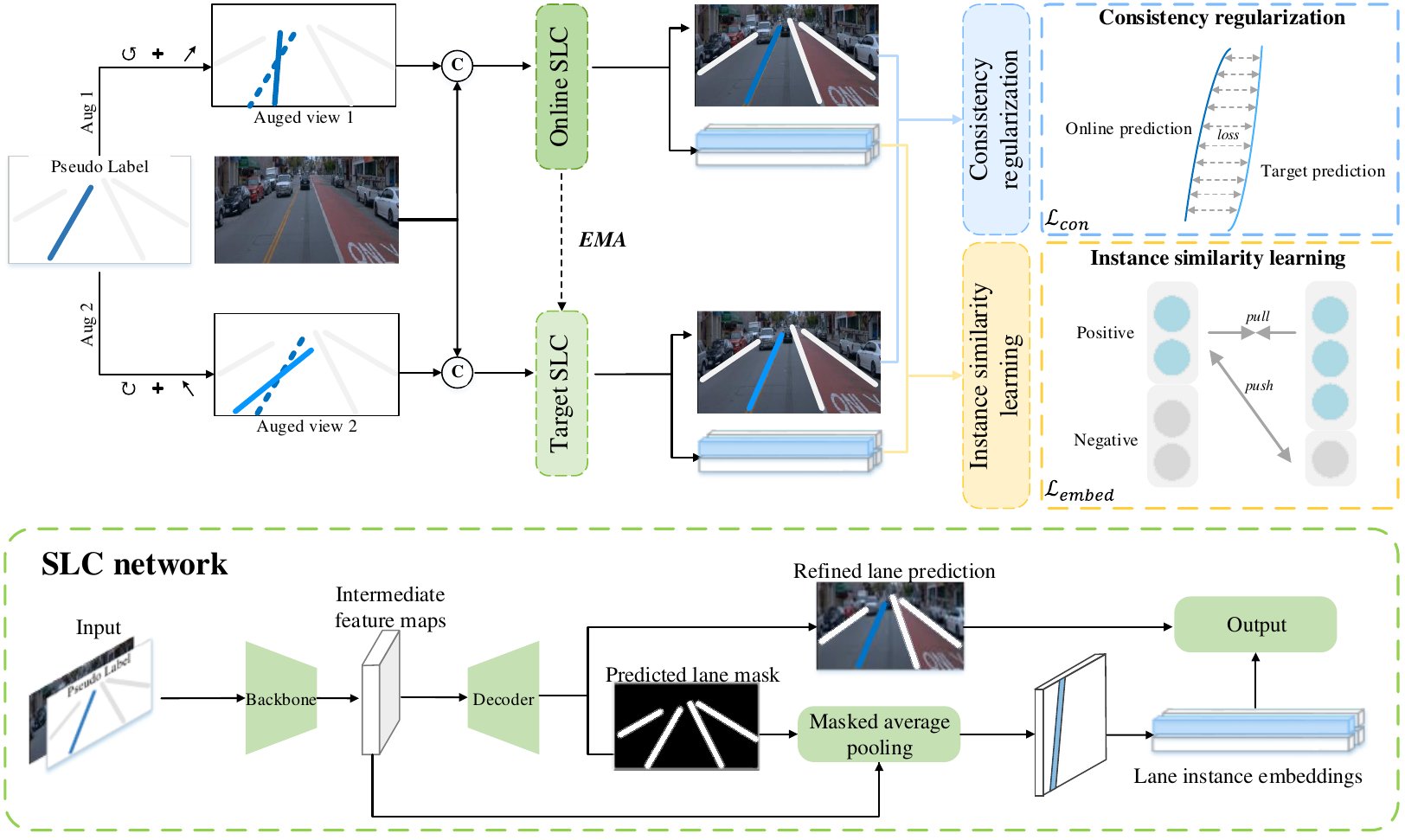}
    \caption{
    Best viewed in color and lane instance number.
    Our \textit{LaneCorrect} consists of two collaborative networks, namely online SLC (updated by gradient descend) and target SLC (updated by moving average), to consistently correct the noisy annotation.
    During training, two different augmented views of pseudo lane annotations are concatenated with images and fed into online and target branches, of which outputs are collected for consistency regularization and instance similarity learning.
    In consistency regularization, predicted lanes of two branches are constrained to map to the unique noise-free lane locations.
    In instance similarity learning, multi-objective contrastive learning is adopted to ensure superior lane representation ability.
    During testing, only online SLC is used to predict refined lanes.
    }
    \label{fig:noisy lane correction module}
\end{figure*}
\noindent\textbf{Boosting noisy pseudo labels.}
We propose a self-supervised lane correction (\textit{SLC}) network, to improve the quality of the noisy pseudo labels and boost the representation learning for lane detection.
The schematic illustration of proposed \textit{SLC} network is shown in Figure~\ref{fig:noisy lane correction module}.
The network takes the noisy results of unsupervised lane segmentation as inputs and performs contrastive learning to consistently correct the noisy annotations.
It contains a consistency regularization module and instance similarity learning module.
The details of the two modules are introduced in the following.

\noindent\textbf{Consistency regularization module.}
We denote the noisy pseudo annotation and its counterpart, \ie, the potential ground-truth annotation as $\tilde{y}$ and $y_{gt}$ respectively, where $\tilde{y}, ~y_{gt} \in \{0,1\}^{H,W}$.
We assume the noise added during the curve fitting and 3D-to-2D projection follows the Gaussian distribution with a stochastic transformation matrix $s$.
Our goal is to learn a function $G$, which can predict the corresponding invertible geometrical transformation matrix $s^T$ given the input image $X \in \mathbb{R}^{3 \times H \times W}$ and the cue of noisy lane labels:

\begin{equation}
    G(X, \tilde{y}) = s^{T} ~~\text{and} ~~y_{gt} = \tilde{y} \cdot s^{T}.
\end{equation}

In view of the self-supervised learning, we utilize the unlabeled data and noisy annotations to enforce the trained model to be line with \textbf{geometry consistency assumption}.
That is, if a realistic geometrical perturbation was to be applied to the pseudo lane annotation, the predicted inverted transformation should not change significantly.
So the multiple disturbances of the same noise-free annotation must be restored to the consistent spatial location in an adversarial manner.
In the light of this, we restrict our model to generate consistent predictions for the input with different slightly geometric perturbation.
Therefore, we design a dual-student framework to learn to capture the geometrical invariability and reconstruct the potential noise-free lane labels.

Following~\cite{Chen2019AutoCorrectDI,map-repair}, a data augmentation method is used for noisy lane annotations.
A pair of lane instances augmented with different geometry noises will be generated by applying rotation and translation operations to the single lane annotation.
To create supervision for missed and mislabelled situations, some lane instances will also be randomly removed or injected.

Let $G$ represent our \textit{SLC} network that consists of a backbone network and a lane prediction head.
The network $G$ can be any kind of existing lane detection model, which shows that our framework is highly adaptable.
$G$ is designed to predict the corrected lane annotations given images and corresponding noisy annotations.
To utilize the pseudo annotation as clue, we augment the original pseudo annotations with symmetrical noise and encode the augmented labels as a binary mask $m \in \{0,1\}^{H,W}$.
With the RGB image inputs $X$ and the guidance of perturbed annotations mask $m$, we hope our network $G$ can reconstruct the noise-free lane predictions, which can be written in a general form:
\begin{equation}\label{2}
\hat{y} = G(X,~m).
\end{equation}

Since ground truth annotation is inaccessible in unsupervised setting, we propose a consistency regularization method to reconstruct the single unique lane instance given annotation clues which have been augmented in two different manners.
In detail, taking two augmented noisy annotation-clue masks $m_1$ and $m_2$ as well as the RGB image $X$ as inputs, our \textit{SLC} model is expected to generate two sets of corrected lane instances $\hat{y}_1$, $\hat{y}_2$:
\begin{equation}
\left\{
\begin{aligned}
&\hat{y}_1 = G_{\theta}(X, ~m_1); \\
&\hat{y}_2 = G_{\phi}(X, ~m_2),\\
\end{aligned}
\right.
\end{equation}
where $G_{\theta}$ is online network updated with gradient, and $G_{\phi}$ is the target network that has the same architecture as the online network.
The parameters of target network $G_{\phi}$ are an exponential moving average of those of the online network $G_{\theta}$.

These two sets of corrected lane predictions refer to the same unique ground truth lane annotations on one single image.
As a result, corrected lane predictions of the online branch of \textit{SLC} can be used to provide supervision for the target branch, which can be written as a consistency loss term:
\begin{equation}\label{3}
\mathcal{L}_{c} = \mathcal{L}_{lane}(\hat{y}_1, ~\hat{y}_2).
\end{equation}
$\mathcal{L}_{lane}$ refers to a general form of lane prediction loss function, which can be specified as any kind of lane detector.

Given merely consistency regularization loss, the mo- \
del will converge to the trivial solution (\eg, generates all-zero lane instance). In order to prevent model collapse, we introduce a reconstruction loss $\mathcal{L}_r$, 
which enables original pseudo annotation $y$ also being utilized to supervise the online network:

\begin{equation}
    \mathcal{L}_r = \mathcal{L}_{lane}(\hat{y}_1, ~y).
\end{equation}
Given regularization loss and reconstruction loss above, our whole consistency regularization loss can be defined as:
\begin{equation}\label{10}
\mathcal{L}_{con} = \mathcal{L}_{c} + \lambda_r\mathcal{L}_{r}.
\end{equation}

$\lambda_r$ denotes the penalty term for reconstruction loss, which is set to 1 in the early stage of training.

Note that although reconstruction loss does help the model converge in the early training and avoid model collapse, it should be adjusted after $\epsilon$ epochs since we want the \textit{SLC} network to output noise-free ground truth annotation instead of original pseudo annotation.
An adjust strategy for $\lambda_r$ is defined as:
\begin{equation}\label{11}
\lambda_r=\left\{
\begin{aligned}
0, \quad &min(IoU(\hat{y}_1, ~y), ~IoU(\hat{y}_2, ~y)) \leq 0.5,\\
1, \quad &otherwise.
\end{aligned}
\right.
\end{equation}
The $IoU$ represents the intersection-of-union score in pixel level.
When the prediction results of our \textit{SLC} network deviate considerably from the given pseudo annotation input $y$, $y$ is supposed to contain some noise and then reconstruction loss is rejected by setting $\lambda_r$ equal to $0$.
Otherwise, $\lambda_r$ is set to 1.
In our training process, we set $\epsilon$ to $10$.

\noindent\textbf{Instance similarity learning module.}
In the above part, the consistency regularization module is utilized to realize the constraint of the noisy annotations.
However, the consistency supervision mentioned above concentrates on the object level, and no effective supervision is provided on the representation level.
To address this issue, we introduce an instance similarity learning module in this part to exploit the appearance similarity on representation level and leverage the trained features to perform label correction.

We consider all the lanes as positive samples and others negative.
By maximizing the representation similarity among different lane instances across online and target branches, further regularization is imposed on our \textit{SLC} network.

After the refined lane annotation $\hat{y}_1$ is predicted by the network $G$, a set of corresponding pooling masks $\{p_1| p_1\in \{0, 1\}^{H_f \times W_f}\}$ of the online branch is encoded according to the predicted lane instance locations.
$p_1$ is the same width and height as the output feature map $f \in \mathbb{R}^{C_f \times H_f \times W_f}$ of the network $G's$ backbone and represents positive samples of the online branch.
Similarly, corresponding target masks $\{p_2\}$ can also be obtained for the target branch.

Then we select the no-lane locations around the predicted lane and encode the negative masks $\{n\}$.

For every positive mask $p$, we downsample the feature map $f$ with the average pooling operation:

\begin{equation}\label{9}
v_p = \frac{1}{\sum_{i,j}p_{i,j}}p \cdot f, \quad v_p \in \mathbb{R}^{C_f}.
\end{equation}

The same calculation is operated on negative masks to get negative vectors $v_n \in \mathbb{R}^{C_f}$.
We then transform each of these vectors with a two-layer MLP, yielding non-linear projections $z_{p_1}, z_{p_2}, z_n \in \mathbb{R}^d$.
Now we introduce the embedding loss function:
\begin{equation}\label{10}
\mathcal{L}_{embed} = log [1 + \sum_{p_2}\sum_{n}exp(z_{p_1}\cdot z_{n} - z_{p_1}\cdot z_{p_2})].
\end{equation}

The multi-target positive samples $\{z_{p_2}\}$ are pooled from lane masks $\{p_2\}$ generated by the target branch in the whole batch.

Together, the noisy label correction module objective can be defined as:
\begin{equation}\label{11}
\mathcal{L} = \mathcal{L}_{con} + \lambda_{embed} \mathcal{L}_{embed},
\end{equation}
where $\lambda_{embed}$ denotes the penalty term for embedding loss.
With the help of instance similarity learning module, the \textit{SLC} network can shape lane instance clusters by inheriting advantages of metric learning and leverage the trained features from self-supervised tasks in lane reconstruction.

\begin{figure}[t]
    \centering
    \includegraphics[scale=0.45]{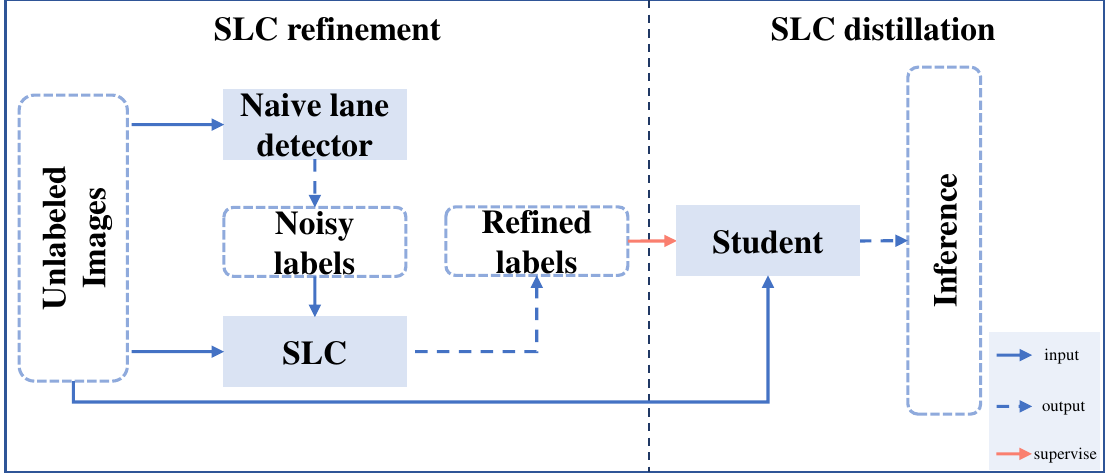}
    \caption{
    The pipeline of pseudo label refinement and distillation.
    To enable our \textit{SLC} to end-to-end inference and better align downstream datasets, we propose a pseudo-label refinement approach in the form of distillation.
    }
    \label{fig:refinement}
\end{figure}

\subsection{Pseudo-label refinement and other details}
As noticed, the \textit{SLC} network accepts both RGB images and pseudo label clues as input.
Thus, although \textit{SLC} predicts noise-free lane detection results, it is infeasible for network to perform further end-to-end inference on downstream datasets.
To alleviate this dilemma, we propose a distillation approach to promote our \textit{SLC} network, as is illustrated in Figure~\ref{fig:refinement}.
In our pseudo-label refinement stage, pseudo annotations are generated by a \textit{na\"ive} lane detector that has been trained on the source dataset as well as our \textit{SLC} network (see Figure~\ref{fig:refinement}).
The noisy results of unsupervised lane segmentation are replaced by these generated pseudo annotation, thus the 3D LiDAR data is not required for further downstream inference.
The corrected predictions of the \textit{SLC} model are used as supervision to train a student lane detector.
The distilled student is capable to conduct end-to-end inference.

Moreover, this procedure can be easily migrated to downstream datasets.
For specifically, we transfer both the \textit{na\"ive} lane detector and \textit{SLC} onto downstream data- \
sets.
The \textit{na\"ive} lane detector generates noisy but cheap annotations for the \textit{SLC}, which \textit{SLC} takes as input and predicts refined lane labels for further distillation.
Finally, the distilled student lane detector is evaluated on the target dataset.

During the whole adaptation process, no supervised pre-trained models are required.
In fact, for each dataset, there is one lane detector training from scratch.

\section{Experiments}
\subsection{Experimental setting}
\noindent\textbf{Datasets.} 
\begin{table}[t]
\renewcommand\arraystretch{1.5}
\caption{Comparisons of the datasets used between supervised paradigm and our method. \cmark and \xmark ~represent whether labels are demanded during training.}
\label{tab:Datasets}
\vspace{-10pt}
\begin{center}
\resizebox{1.0\linewidth}{0.06\textheight}{
\begin{tabular}{lccccc}
\midrule[1.0pt]
\rowcolor[gray]{0.8}& \multicolumn{2}{c}{Pre-train} & \multicolumn{2}{c}{Train} & Test \\
\rowcolor[gray]{0.8}\multirow{-2}{*}{Method}& dataset & label & dataset & label & dataset \\
\midrule[1.0pt]
Supervised & ImageNet & \cmark & \ie TuSimple & \cmark & \ie TuSimple \\
Ours & Waymo & \xmark & \ie TuSimple & \xmark & \ie TuSimple \\
\midrule[1.0pt]
\end{tabular}}
\end{center}
\vspace{-15pt}
\end{table}
We pre-train our \textit{LaneCorrect} on large-scale \textit{Waymo Open} dataset~\citep{sun2020scalability}.
It contains synchronous LiDAR frames and 2D images.
To evaluate our proposed method and make fair comparisons, we distill the \textit{SLC} network on the target domain to train a student lane detector which can be evaluated directly on the testing set.
With the help of our self-supervised pre-trained \textit{SLC} model, the domain gap is alleviated.
Note that we do not introduce LiDAR data in the inference phase and no ImageNet pre-trained model is included in our method.
Only the supervised counterparts use backbones pre-trained on ImageNet for comparison.
The details can be viewed in Table~\ref{tab:Datasets}.

We conduct evaluations on four datasets: \textit{TuSimple}~\cite{tusimple}, \textit{CULane}~\cite{pan2018spatial}, \textit{LLAMAS}~\cite{llamas2019} and \textit{CurveLanes}~\cite{CurveLane-NAS}.
\textbf{\textit{TuSimple:}} \textit{TuSimple} is a widely used dataset targeted to solve the lane detection problem on highways. 
It includes 3626 training video clips and 2782 test video clips.
Only good weather conditions and daytime data are given.
\textbf{\textit{CULane:}} \textit{CULane} is a traffic lane detection dataset collected in Beijing and released by Chinese University of Hong Kong. 
This dataset provides 133,235 frames form a 55 hours of videos, which is then be divided into training, validation and test set by 88880, 9675 and 34680. 
Specially, the test set provides abundant scene, including one normal and 8 challenging categories.
\textbf{\textit{LLAMAS:}} \textit{LLAMAS} is one of the largest lane detection datasets with over 100,000 images with highway scenarios.
The annotations of \textit{LLAMAS} are automatically generated by the high definition map projections.
Its test set evaluation is provided by the official \textit{LLAMAS} server.
\textbf{\textit{CurveLanes:}} \textit{CurveLanes} contains 150K images with human annotated lane labels for difficult scenarios in traffic lane detection.
In some scenarios of this dataset, the lane detection task is quite complex and challenging due to curve and fork lanes.
The details of the datasets can be viewed in Table \ref{tab:Datasets overview}.

\begin{table}[t]
\renewcommand\arraystretch{1.5}
\caption{Details of lane detection datasets utilized.}
\label{tab:Datasets overview}
\vspace{-15pt}
\begin{center}
\resizebox{1.0\linewidth}{0.08\textheight}{
\begin{tabular}{lcccc}
\midrule[1.0pt]
\rowcolor[gray]{0.8}Dataset  & Train & Val. & Test & Scenario Type \\ 
\midrule[1.0pt]
TuSimple & 3k    & 0.3k   &   2k  & Highway      \\
CULane &  88k    & 9k  &  34k  & Urban\&Highway     \\
LLAMAS   & 58k    & 20k  &  20k & Highway      \\
CurveLanes  & 100k  & 20k & 30k & Urban\&Highway  \\
\midrule[1.0pt]
\end{tabular}}
\end{center}
\vspace{-15pt}
\end{table}

\noindent\textbf{Implementation details.} To generate pseudo 3D lane segmented points and corresponding 2D coordinates, we firstly perform segmentation algorithm on \textit{Waymo Open} dataset~\cite{sun2020scalability}.
We only select synchronized 2D images collected by the camera in the front direction and 3D point clouds from the top LiDAR as our inputs.
We choose PointLaneNet~\cite{pointlanenet} as our \textit{na\"ive} lane detection network and our supervised baseline, with ResNet101~\cite{resnet} as backbone in the main results.
We additionally run experiments with LaneATT~\cite{Tabelini2020KeepYE} as our baseline in ablation to better prove our framework adaptable to arbitrary lane detectors.
During \textit{SLC} network training, the augmentation method consists of random rotation between $-5^\circ$ and $5^\circ$ and pixel translation of up to $5\%$ of the width of the input image.
We apply an adjusting strategy for reconstruction loss weight to prevent the model overfit to pseudo annotations, which has been detailed described in consistency regularization module.
The hyper-paremeter $\lambda_{embed}$ is set to $5$ according to ablation study.
All other hyper-parameter settings follow PointLaneNet~\cite{pointlanenet}, and our whole architecture is conducted on 8 Nvidia V100 GPU cards.

\noindent\textbf{Details of base lane detector.}
In our experiments, we adopt the anchor-based counterpart PointLaneNet \
\cite{pointlanenet} as the \textit{na\"ive} lane detector ($\mathcal{L}_{lane}$ in Eq~4).
PointLaneNet~\cite{pointlanenet} is able to simultaneously perform position prediction and lane classification in a single network.
Two $1 \times 1$ convolution layers are added on the top of the backbone network, specifying the number of output channels equal to $(n+1)$, where $n$ refers to the number of $x$ coordinate offsets $\{\delta x_1, \delta x_2, ..., \delta x_n\}$ (fixed $y$ partitions) relative to the center point of the grid, and $1$ refers to the starting position $(y)$ of the lane.
As for classification, after two $1 \times 1$ convolution layers to the end of the feature map, the number of output channels is specified equal to 2, indicating whether the lane passes through the grid.
Given ground truth $y_{gt}$, the objective is to optimize the multi-part loss:

\begin{equation}\label{1}
\resizebox{0.34\textwidth}{!}{$
\begin{aligned}
\mathcal{L}_{lane}(y,~y_{gt}) = 
\frac{1}{N_{cls}}\sum_{i=1}^{w}\sum_{j=1}^{h}\mathcal{L}_{ij}^{cls}(y_{ij}^{cls},~y^{cls}_{ij,~gt}) \\
+ \frac{1}{N_{reg}}\sum_{i=1}^{w}\sum_{j=1}^{h}l_{ij}^{obj}\mathcal{L}_{ij}^{reg}(y_{ij}^{reg},~y^{reg}_{ij,~gt}).
\end{aligned}$}
\end{equation}
$\mathcal{L}_{ij}^{cls}(y_{ij}^{cls},~y^{cls}_{ij,~gt})$ represents the classification confidence loss at anchor $(i,~j)$, which is cross-entropy loss between the prediction results and ground truth.
For each anchor with the classification prediction $y^{cls}$ and $y_{gt}^{cls}$, the confidence loss is written as:

\begin{equation}\label{2}
\begin{aligned}
&\mathcal{L}^{cls}(y^{cls}, y_{gt}^{cls}) = \\
&{-\left[y_{gt}^{cls}lny^{cls} + (1-y_{gt}^{cls})ln(1-y^{cls}) \right]}.
\end{aligned}
\end{equation}

$\mathcal{L}_{ij}^{reg}(y_{ij}^{reg},~ y^{reg}_{ij,~gt})$ denotes the Euclidean distance between the predicted locations and ground truth at anchor $(i,~j)$.
In general, each anchor generates the regression prediction $y^{reg}=\{y_1^{reg}, y_2^{reg}, ..., y_n^{reg}, y_{pos}^{reg}\}$, where $y_1^{reg}$, $y_2^{reg}$, ..., $y_n^{reg}$ represent the $n$ offsets prediction and $y_{pos}^{reg}$ represents the starting position of the lane.
The loss term can be written as:

\begin{equation}\label{3}
\begin{aligned}
\mathcal{L}^{reg}(y^{reg},~y^{reg}_{gt}) = 
\sum_{k=1}^{n}(y_{k}^{reg}-y_{k,~gt}^{reg})^{2}.
\end{aligned}
\end{equation}

In our proposed \textit{LaneCorrect}, we adopt the above lane prediction loss function to formulate our reconstruction loss $\mathcal{L}_r$ and consistency loss $\mathcal{L}_c$ in our consistency regularization module:
\begin{equation}\label{4}
\begin{aligned}
\mathcal{L}_{c}(\hat{y}_1,~\hat{y}_2) =
\mathcal{L}_{lane}(\hat{y}_1,~\hat{y}_2),
\end{aligned}
\end{equation}
and
\begin{equation}\label{4}
\begin{aligned}
\mathcal{L}_{r}(\hat{y}_1,~\hat{y}_2) =
\mathcal{L}_{lane}(\hat{y}_1,~y),
\end{aligned}
\end{equation}
where $\hat{y}_1$ and $\hat{y}_2$ are lane predictions of online branch and target branch, and $y$ denotes input pseudo annotations.

\begin{table}[t]
\renewcommand\arraystretch{1.5}
\caption{Performance of the proposed method and comparison with counterpart on \textit{TuSimple} testing set. Supervised PointLaneNet is trained directly using manual annotations. Ours (Waymo pre-trained) is proposed \textit{na\"ive} unsupervised lane detector. Ours (\textit{SLC}) is further applied noisy lane correction network. Our method achieves considerable performance.}
\label{tab:tusimple result}
\vspace{-10pt}
\begin{center}
\resizebox{1.0\linewidth}{0.06\textheight}{
\begin{tabular}{lcccc}
\midrule[1.0pt]
\rowcolor[gray]{0.8}Method  & F1(\%)$\uparrow$ & ACC(\%)$\uparrow$ & FPR(\%)$\downarrow$ & FNR(\%)$\downarrow$\\ 
\midrule[1.0pt]
Supervised (PointLaneNet) & 95.07 & 96.34 & 4.67 & 5.18 \\
\midrule
Ours (Waymo pre-trained) & 83.35 & 89.34 & 18.02 & 12.63 \\
Ours (\textit{SLC}) & 92.91 & 91.95 & 6.45 & 6.58\\
\midrule[1.0pt]
\end{tabular}}
\end{center}
\vspace{-10pt}
\end{table}

\begin{table}[t]
\renewcommand\arraystretch{1.5}
\caption{Comparison with comparative learning-based counterpart on \textit{TuSimple} testing set.
The first column indicates the proportion of labeled data used for training.
The efficiency of our method towards data utilization is more pronounced.}
\label{tab:erfnet}
\vspace{-10pt}
\begin{center}
\resizebox{1.0\linewidth}{0.14\textheight}{
\begin{tabular}{lccc}
\midrule[1.0pt]
\rowcolor[gray]{0.8}Labels & Pseudo labels & Methods & ACC(\%)$\uparrow$ \\
\midrule[1.0pt]
100\% & - & ERFNet~\cite{lin2021semi} & 93.71 \\
\midrule
50\% & Hough Transform & s4GAN~\cite{mittal2019s4} & 88.82 \\
50\% & Hough Transform & ERFNet & 93.70 \\
50\% & 3D lane segmentation & Ours & \textbf{94.04} \\
\midrule
10\% & Hough Transform & s4GAN & 86.25 \\
10\% & Hough Transform & ERFNet & 93.05 \\
10\% & 3D lane segmentation & Ours & \textbf{93.58} \\
\midrule
0\% & Hough Transform & s4GAN & - \\
0\% & Hough Transform & ERFNet & - \\
0\% & 3D lane segmentation & Ours & \textbf{91.95} \\
\midrule[1.0pt]
\end{tabular}}
\end{center}
\vspace{-10pt}
\end{table}

\begin{table*}[ht]
\caption{Comparison of F1-measure on \textit{CULane} testing set. For the \textit{Cross} scenario, only false positives are shown. The less number means the better performance. Our \textit{SLC} network can significantly alleviate the domain gap compared with our unsupervised \textit{na\"ive} lane detector.}
\label{tab:culane result}
\vspace{-10pt}
	\centering
        \scalebox{0.93}{
	\setlength{\tabcolsep}{5pt} 
	\renewcommand{\arraystretch}{1} 
	\setlength{\tabcolsep}{1mm}{
		\begin{tabular}{lcccccccccc}
		    \midrule[1.0pt]
			\rowcolor[gray]{0.8}Method     & Total & Normal  & Crowded & Dazzle & Shadow  & No Line  & Arrow & Curve & Cross & Night\\ 
			\midrule[1.0pt]
			Supervised (PointLaneNet) & 70.2 & 88.0 & 68.1  & 61.5 & 63.3 & 44.0 & 80.9 & 65.2 & 1640 & 63.2  \\
		    \midrule
		    Ours (Waymo pre-trained) & 46.8 & 57.7 & 40.9 & 41.5 & 35.3 & 32.4 & 51.3 & 48.3 & 3210 & 41.5\\
		    Ours (\textit{SLC}) & 56.7 & 69.9 & 57.1 & 52.5 & 49.0 & 38.4 & 68.4 & 53.7 & 2385 & 49.0\\
		    \midrule[1.0pt]
	\end{tabular}}}
	\vspace{-15pt}
\end{table*}

\subsection{Main Results}
Since there was no previous work addressing unsupervised lane detection task, to conduct comparison fairly, we directly compare our methods with supervised PointLaneNet~\cite{pointlanenet} as counterpart.
The comparison shows that our self-supervised method achieves competitive results compared with supervised methods.
On \textit{LLAMAS} val. set, our method even outperforms its supervised counterpart.

\noindent\textbf{Ablated effects of pre-training on Waymo.}
To make fair comparison and address the concern of utilizing extra \textit{Waymo}~\cite{sun2020scalability} dataset, we both conduct experiments with and without \textit{SLC} network from Table~\ref{tab:tusimple result} to Table~\ref{tab:curvelanes result}.
The performance of model without \textit{SLC} network, \ie, \textit{na\"ive} lane detector simply trained with pseudo labels from \textit{Waymo}, denotes the potential effects of pre-training on \textit{Waymo} dataset.
Additionally, the \textit{SLC} network brings significant improvement to the \textit{na\"ive} lane detector, demonstrating the effectiveness of the self-supervised lane detection framework.

\noindent\textbf{TuSimple.} For \textit{TuSimple} benchmark, we report both accuracy and F1 score in Table \ref{tab:tusimple result}.
As demonstrated, the gap between method of strongly supervised and our method is quitely small.
Especially, our self-supervised method achieves considerable results at $92.91\%$ F1-mea- \
sure compared with supervised counterpart, which achie- \
ves $95.07\%$ at F1-measure.
The comparison result demonstrates the efficiency of our proposed framework.

We also compare our method with other comparative learning-based approaches in Table~\ref{tab:erfnet}.
The first column indicates the proportion of labeled data used for training, while the remaining data is treated as unlabeled for self or semi-supervised learning.
Notably, our method is capable of handling scenarios when there is no (0\%) human-labeled data included.
Additionally, the efficiency of our method is more pronounced across all situations.

\noindent\textbf{CULane.}
The results on \textit{CULane} can be seen in Table \ref{tab:culane result}.
Our method achieves $55.7\%$ F1-score in total.
We observed that our LaneCorrect model encounters a more pronounced domain gap on CULane (collected in Beijing) compared to TuSimple (collected in the United States, similar to Waymo).
Despite the significant domain disparity between our self-supervised source domain and the target CULane dataset, our unsupervised method with the SLC network achieves a reasonable F1-score compared to its strongly supervised counterpart, without touching any ground truth.
The comparison result also demonstrates the \textit{SLC} network brings significant improvements to our unsupervised method and can effectively alleviate the domain gap.

\begin{figure}[h!]
    \centering
    \includegraphics[scale=0.31]{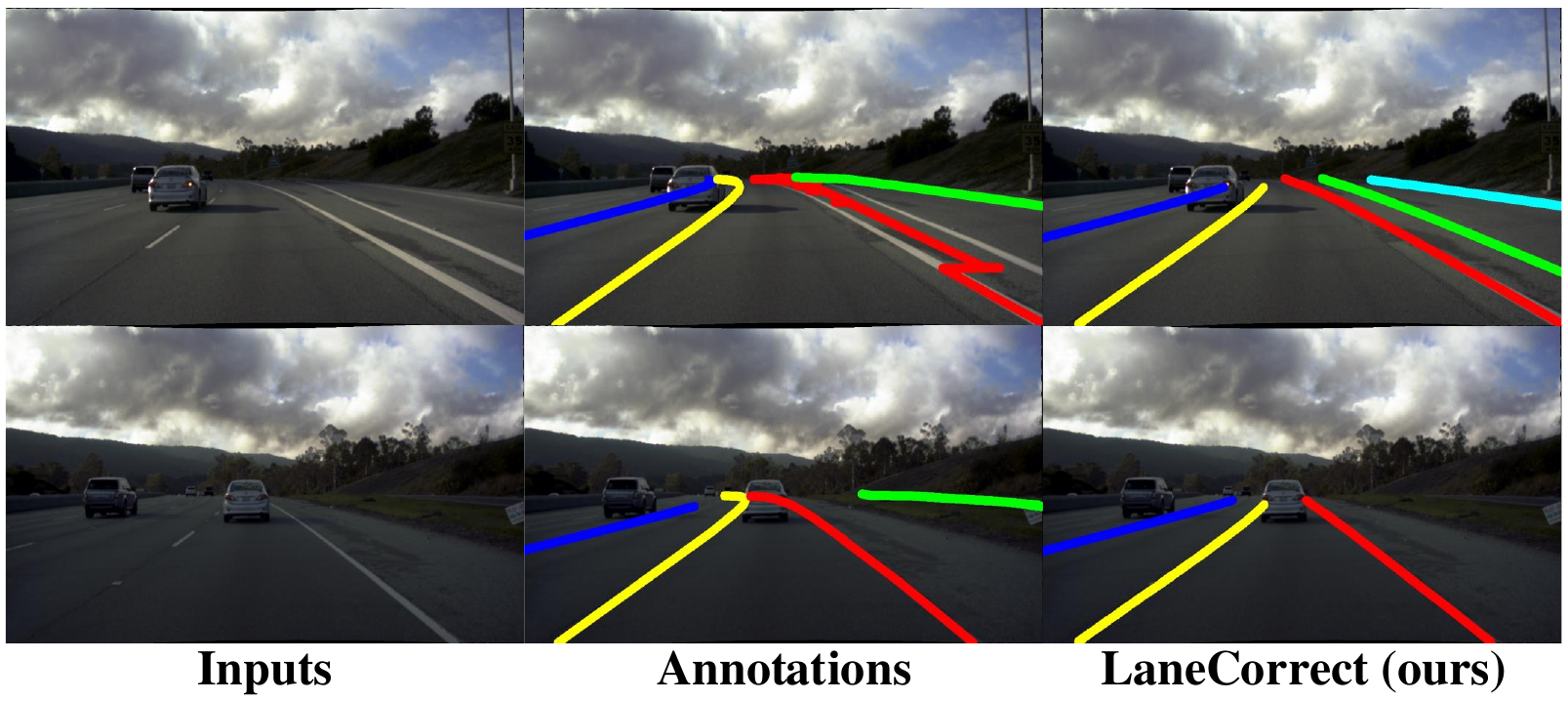}
    \caption{Incorrect annotated cases in \textit{LLAMAS}.
    Our method can generate correct predictions in these incorrect labeled scenarios.}
    \label{fig:incorrect annotated cases}
\end{figure}

\noindent\textbf{LLAMAS.}
The performance on \textit{LLAMAS} is shown in Table \ref{tab:llamas all}.
Our model is trained only with \textit{LLAMAS} training images and is evaluated on validation set and official test set.
As the results demonstrate,  our unsupervised method achieves considerable results at $89.75\%$ F1-measure at testing set and $91.29\%$ at validation set.
It is worth noting that, our unsupervised lane detection framework outperforms supervised method on \textit{LLAMAS} validation set.
One of the main reason is that \textit{LLAMAS} dataset is an unsupervised dataset annotated automatically by high-resolution maps and its projection.
Due to uncertainty error and projection bias, the annotations generated sometimes are incorrect, which will damage the performance of the lane detector.
The incorrect case is also visualized in Figure \ref{fig:incorrect annotated cases}.
From the visualization result, we can see that our unsupervised method is able to perform well in mislabeled scenarios.

\begin{figure*}[th]
\centering
\includegraphics[scale=0.9]{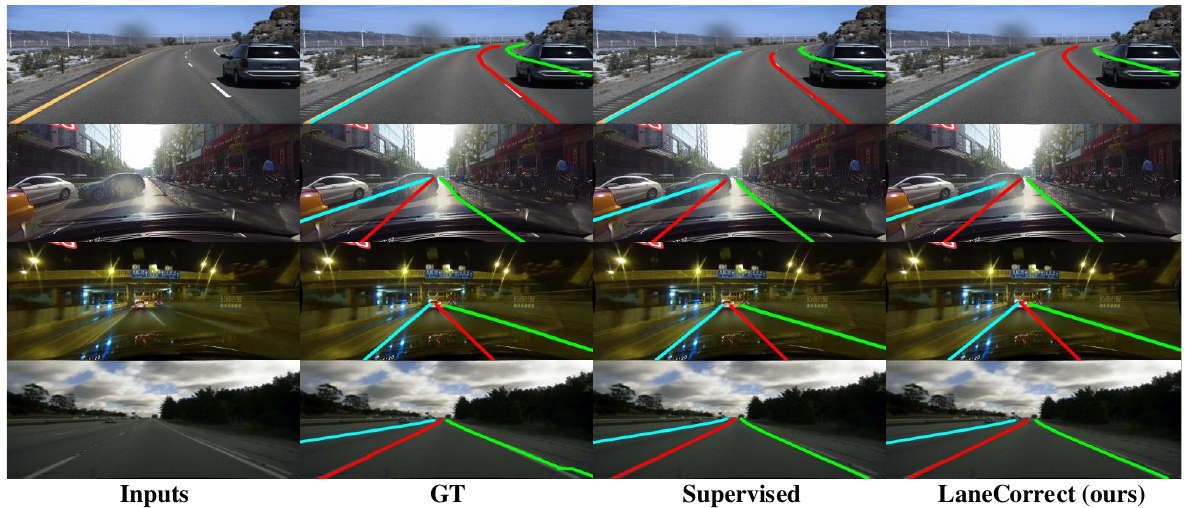}
\caption{Visualization of \textit{LaneCorrect} method on multiple benchmarks compared with supervised counterpart. The top row is performance on \textit{TuSimple} and the bottom row is performance on \textit{LLAMAS}. The rest middle rows are qualitative results on \textit{CULane}. For each row, from left to right are: input image, ground truth, results of supervised counterpart and our \textit{LaneCorrect}.}
\label{fig:main results}
\end{figure*}

\begin{table}[t]
\renewcommand\arraystretch{1.5}
\caption{Comparison on \textit{LLAMAS} validation and testing set. Our method receives considerable results and even outperforms supervised method on validation set.}
\label{tab:llamas all}
\vspace{-10pt}
\begin{center}
\resizebox{1.0\linewidth}{0.09\textheight}{
\begin{tabular}{lcccc}
\midrule[1.0pt]
\rowcolor[gray]{0.8}Dataset & Method  & F1(\%)$\uparrow$ & Precision(\%)$\uparrow$ & Recall(\%)$\uparrow$\\ 
\midrule[1.0pt]
\multirow{3}{*}{Val.} & Supervised (PointLaneNet) & 90.32 & 96.51 & 84.87\\
& Ours (Waymo pre-trained) & 87.44 & 89.13 & 85.81\\
& Ours (\textit{SLC}) & \textbf{91.29} & 92.47  & \textbf{90.13}\\
\midrule
\multirow{3}{*}{Test} & Supervised (PointLaneNet) & 95.11 & 95.17 & 95.05\\
& Ours (Waymo pre-trained) & 85.71 & 85.24 & 86.18\\
& Ours (\textit{SLC}) & 89.05 & 93.95  & 84.64\\
\midrule[1.0pt]
\end{tabular}}
\end{center}
\vspace{-10pt}
\end{table}
\begin{table}[t]
\renewcommand\arraystretch{1.5}
\caption{Comparison with counterpart on \textit{CurveLanes} testing set. Our method shows reasonable result and domain gap has been mitigated by the proposed \textit{SLC} network.}
\label{tab:curvelanes result}
\vspace{-10pt}
\begin{center}
\resizebox{1.0\linewidth}{0.07\textheight}{
\begin{tabular}{lcccc}
\midrule[1.0pt]
\rowcolor[gray]{0.8}Method  & F1(\%)$\uparrow$ & Precision(\%)$\uparrow$ & Recall(\%)$\uparrow$\\ 
\midrule[1.0pt]
Supervised (PointLaneNet) & 78.47 & 86.33 & 72.91\\
\midrule
Ours (Waymo pre-trained) & 49.24 & 68.71 & 38.37\\
Ours (\textit{SLC}) & 60.39 & 62.67 & 58.27\\
\midrule[1.0pt]
\end{tabular}}
\end{center}
\vspace{-10pt}
\end{table}
\noindent\textbf{CurveLanes.}
Table \ref{tab:curvelanes result} shows the performance of our method on \textit{CurveLanes}.
As the results show, our \textit{LaneCorrect} achieves considerable results at $60.39\%$ F1-measure, which is a reasonable result compared with supervised method.
The visualization results are shown in Fig.~\ref{fig:main results}.

\subsection{Ablation study}
In this part of experiment, we evaluate the impact of the major components of our self-supervised \textit{SLC} model and the variation of other experimental settings.
The ablation study is performed on \textit{TuSimple} dataset.

\noindent\textbf{Modules of noisy lane correction network.}
During training, the \textit{SLC} network improve the unsupervised \textit{na\"ive} lane detector significantly.
In this work, we explore the benefits gained from each part of proposed network.
Table \ref{tab:ablation study} shows that the performance is constantly improved with the gradual introduction of reconstruction loss, consistency regularization and instance similarity learning module.
For experiment with reconstruction loss, only the online branch of the \textit{SLC} network is used.
For experiment with consistency regularization, both the branches of the network are used but the embedding head for instance similarity learning is removed.
Experiment with contrastive instance similarity learning module is the full version of our \textit{LaneCorrect} method, which achieves $92.91\%$ F1-score at \textit{TuSimple}.

\begin{table}[t]
\renewcommand\arraystretch{1.5}
\caption{Quantitative results of ablation study of our self-supervised noisy lane correction network on \textit{TuSimple}. As different portions of the proposed \textit{SLC} network introduced, the gap between our unsupervised method and supervised counterpart is gradually reduced.}
\label{tab:ablation study}
\vspace{-10pt}
\begin{center}
\resizebox{1.0\linewidth}{0.07\textheight}{
\begin{tabular}{lcccc}
\midrule[1.0pt]
\rowcolor[gray]{0.8}Method  & F1(\%)$\uparrow$ & ACC(\%)$\uparrow$ & FPR(\%)$\downarrow$ & FNR(\%)$\downarrow$\\ 
\midrule[1.0pt]
Supervised (PointLaneNet) & 95.07 & 96.34 & 4.67 & 5.18 \\
\midrule
\textit{Unsupervised na\"ive lane detector} & 84.08 & 85.97 & 13.82 & 14.10 \\
+ reconstruction loss & $88.75^{\textbf{+4.67}}$ & $87.82^{\textbf{+1.85}}$ & $9.50^{\textbf{-4.32}}$ & $10.36^{\textbf{-3.74}}$ \\
+ consistency regularization & $91.89^{\textbf{+3.14}}$ & $90.65^{\textbf{+2.83}}$ & $7.03^{\textbf{-2.47}}$ & $7.56^{\textbf{-2.80}}$ \\
+ instance similarity learning & $92.91^{\textbf{+1.02}}$ & $91.95^{\textbf{+1.30}}$ & $6.45^{\textbf{-0.58}}$ & $6.58^{\textbf{-0.98}}$ \\
\midrule[1.0pt]
\end{tabular}}
\end{center}
\vspace{-10pt}
\end{table}

\begin{figure}[ht]
    \centering
    \includegraphics[scale=0.75]{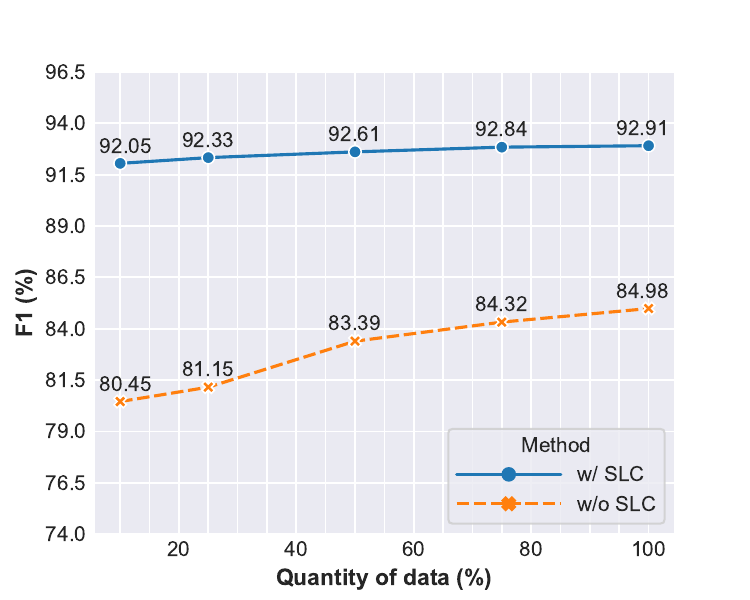}
    \caption{Evaluation results on \textit{TuSimple} when various amount of data are used for pseudo lane annotations generation.}
    \label{fig:data efficiency}
\end{figure}

\noindent\textbf{Efficiency in data utilization.}
To examine the impact of various amount of data used for pre-training, we randomly sample different portions of \textit{Waymo Open} dataset according to the video sequence.
By conducting this experiment, we prove that efficient data utilization of our self-supervised \textit{SLC} method is the main reason why our proposed methods achieve excellent results.

As is shown in Figure \ref{fig:data efficiency}, the performance of both our unsupervised \textit{na\"ive} lane detector and student lane detector distilled by \textit{SLC} network gradually improves as the amount of data used for training increases.
On the other hand, with \textit{SLC} model, we are able to achieve on par performance with a quite small amount of training data, compared with the whole portion of \textit{Waymo Open} dataset.
This further demonstrates the effectiveness of the proposed \textit{SLC} network.

\noindent\textbf{Generalization.}
Inspired by~\cite{Qu2021FocusOL}, we perform experiments on cross domain tasks.
To test the generalization performance of \textit{SLC}, the student lane detector distilled by our \textit{SLC} network on \textit{CULane} domain is evaluated on \textit{TuSimple} testing set.
We also transfer the supervised PointLaneNet model trained on the \textit{CULane} training set to \textit{TuSimple} testing set.
Table \ref{tab:transfer learning result1} shows that compared with supervised method, \textit{LaneCorrect} achieves excellent progress in generalization, especially in terms of FPR and FNR.
Even compared with other state-of-the-art supervised method, our unsupervised framework achieves comparable results, which proves the generalization ability of our method.

\begin{table}[t]
\renewcommand\arraystretch{1.5}
\caption{Comparison about generalization ability with other supervised methods from \textit{CULane} training set to \textit{TuSimple} testing set. ``*" represents that the results are from original paper~\cite{Qu2021FocusOL}. ``-" means that the results are not reported. Our method achieves on par performance with state-of-the-art method at accuracy, but has lower FP and FN rates, resulting a much higher F1 than supervised method.}
\label{tab:transfer learning result1}
\vspace{-10pt}
\begin{center}
\resizebox{1.0\linewidth}{0.1\textheight}{
\begin{tabular}{lcccc}
\midrule[1.0pt]
\rowcolor[gray]{0.8}Method & F1(\%)$\uparrow$ & ACC(\%)$\uparrow$ & FPR(\%)$\downarrow$ & FNR(\%)$\downarrow$ \\ 
\midrule[1.0pt]
SIM-CycleGAN+ERFNet * & - & 62.58 & 98.86 & 99.09 \\
UFNet * & - & 65.53 & 56.80 & 65.46 \\
PINet(4H) * & - & 36.31 & 48.86 & 89.88 \\
FOLOLane * & - & 84.36 & 39.64 & 38.41 \\
\midrule
Supervised (PointLaneNet) & 16.03 & 53.27 & 42.00 & 44.97 \\
\midrule
Ours (Waymo pre-trained) & 22.53 & 58.61 & 38.07 & 36.83 \\
Ours (\textit{SLC})     & \textbf{67.18}& \textbf{85.34} & \textbf{20.11} & \textbf{29.63} \\
\midrule
\end{tabular}}
\end{center}
\vspace{-10pt}
\end{table}

\begin{table}[t]
\renewcommand\arraystretch{1.5}
\caption{Performance of the proposed method and comparison with counterpart on \textit{TuSimple} testing set using LaneATT and CLRNet. To make comparison, we report results using PointLaneNet as baseline.}
\label{tab:tusimple result laneatt}
\vspace{-10pt}
\begin{center}
\resizebox{1.0\linewidth}{0.1\textheight}{
\begin{tabular}{lccccc}
\midrule[1.0pt]
\rowcolor[gray]{0.8}Baseline & Method  & F1(\%)$\uparrow$ & ACC(\%)$\uparrow$ & FPR(\%)$\downarrow$ & FNR(\%)$\downarrow$\\ 
\midrule[1.0pt]
\multirow{3}{*}{PointLaneNet} & Supervised & 95.07 & 96.34 & 4.67 & 5.18 \\
& Waymo pre-trained & 83.35 & 89.34 & 18.02 & 12.63 \\
& Ours (\textit{SLC}) & 92.91 & 91.95 & 6.45 & 6.58\\
\midrule
\multirow{3}{*}{LaneATT} & Supervised & 96.06 & 96.10 & 4.64 & 2.17 \\
& Waymo pre-trained & 86.75 & 90.03 & 14.58 & 9.87 \\
& Ours (\textit{SLC}) & 93.95 & 93.68 & 5.41 & 5.06\\
\midrule
\multirow{3}{*}{CLRNet} & Supervised & 97.62 & 96.83 & 2.37 & 2.38 \\
& Waymo pre-trained & 88.59 & 92.16 & 8.62 & 6.46 \\
& Ours (\textit{SLC}) & 96.91 & 96.06 & 2.57 & 2.72 \\
\midrule[1.0pt]
\end{tabular}}
\end{center}
\vspace{-10pt}
\end{table}
\begin{table}[t]
\renewcommand\arraystretch{1.5}
\caption{Improvements of LaneCorrect algorithm at each step on Waymo.}
\label{tab:stepwise}
\vspace{-10pt}
\begin{center}
\resizebox{1.0\linewidth}{0.05\textheight}{
\begin{tabular}{lcccc}
\midrule[1.0pt]
\rowcolor[gray]{0.8}Method  & F1(\%)$\uparrow$ & ACC(\%)$\uparrow$ & FPR(\%)$\downarrow$ & FNR(\%)$\downarrow$\\ 
\midrule[1.0pt]
Noisy pseudo labels & 79.68 & 85.76 & 15.34 & 13.87 \\
Na\"ive lane detector & 82.75 & 86.53 & 10.82 & 11.21 \\
SLC network & 92.63 & 95.88 & 4.17 & 3.85 \\
\midrule[1.0pt]
\end{tabular}}
\end{center}
\vspace{-10pt}
\end{table}
\begin{table}[t]
\renewcommand\arraystretch{1.5}
\caption{Ablation studies on hyper-parameter $\lambda_{embed}$.}
\label{tab:hyper}
\vspace{-10pt}
\begin{center}
\resizebox{1.0\linewidth}{0.07\textheight}{
\begin{tabular}{lcccc}
\midrule[1.0pt]
\rowcolor[gray]{0.8}$\lambda_{embed}$ & F1(\%)$\uparrow$ & ACC(\%)$\uparrow$ & FPR(\%)$\downarrow$ & FNR(\%)$\downarrow$ \\
\midrule[1.0pt]
1 & 91.05 & 89.47 & 7.86 & 8.12 \\
5 & \textbf{92.91} & \textbf{91.95} & \textbf{6.45} & \textbf{6.58} \\
10 & 91.83 & 90.23 & 7.17 & 7.54\\
\midrule[1.0pt]
\end{tabular}}
\end{center}
\vspace{-10pt}
\end{table}

\begin{figure*}[t]
\includegraphics[scale=0.9]{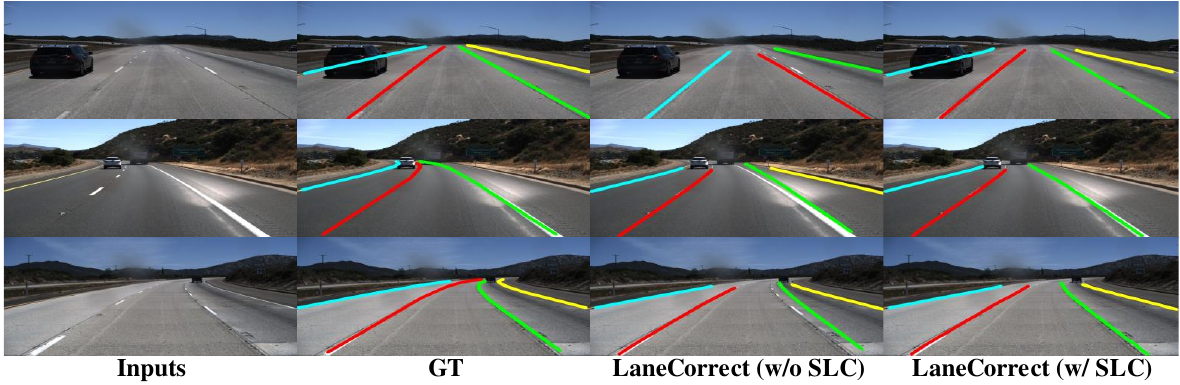}
\caption{Qualitative performance of \textit{SLC} network on \textit{TuSimple} dataset compared with \textit{na\"ive} lane detector. For each row, from left to right are: input image, ground truth, results of unsupervised \textit{na\"ive} lane detector without \textit{SLC} network and results of our \textit{LULA} method with \textit{SLC} network.}
\label{fig:nlc}
\end{figure*}
\begin{figure*}[t]
\includegraphics[scale=0.9]{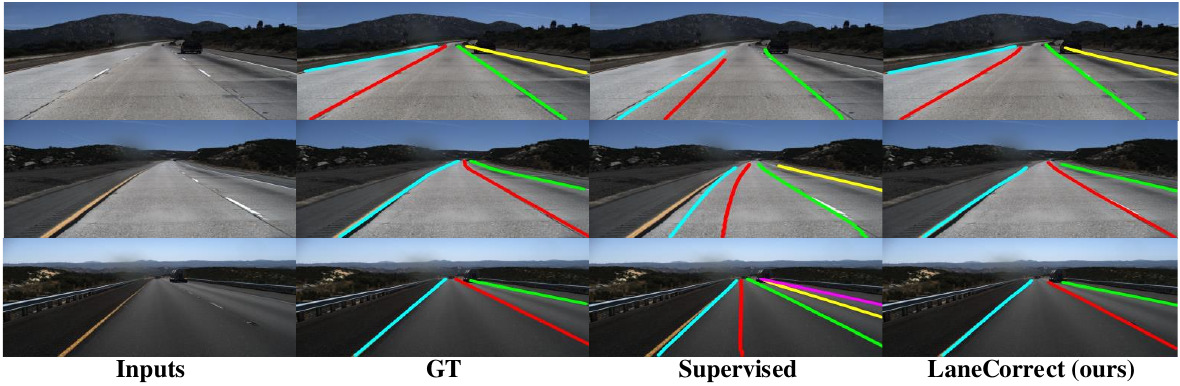}
\caption{Visualization of cross domain performance from \textit{CULane} to \textit{TuSimple}. For each row, from left to right are: input image, ground truth, results of supervised counterpart and our \textit{LaneCorrect}.}
\label{fig:da}
\end{figure*}

\noindent\textbf{Extendability.}
In our main experiments, we choose PointLaneNet as our \textit{na\"ive} lane detector because it is a simple and stable lane detection method.
Our \textit{SLC} network is also based on the PointLaneNet.
The design of online and target branch enables our \textit{SLC} network to realize regularization on geometric consistency and contrast learning in feature representation, so as to correct noisy annotations.
To clarify the robust and extendability, we also carry out experiments using LaneATT \
\cite{Tabelini2020KeepYE} and CLRNet~\cite{zheng2022clrnet} in Table~\ref{tab:tusimple result laneatt}.

The LaneATT and CLRNet baseline achieve considerable results at $93.95\%$ and $96.91\%$ in F1-measure compared with supervised counterpart.
Also as the comparison result demonstrates, when a more efficient supervised baseline is adopted, the performance of our unsupervised baseline algorithm and \textit{SLC} network will also be higher accordingly.

\noindent\textbf{Improvements of SLC network.}
We are also interested in assessing the quality of noisy pseudo annotations and examining the enhancements brought about by the SLC network.
In this section, we conduct experiments to evaluate the pseudo annotations, the na\"ive lane detector, and the SLC network directly on the Waymo dataset, as shown in Table~\ref{tab:stepwise}.
Since Waymo does not provide lane labels, we manually annotate 2000 samples of the Waymo dataset to create a test subset for evaluation.
The results demonstrate the performance of refined pseudo labels generated by the SLC network, thus confirming the high quality of the corrected annotations by our method.

\noindent\textbf{Choice of loss balance term $\lambda_{embed}$.}
We conduct ablation studies on \textit{Tusimple} dataset to determine the best setting of $\lambda_{embed}$.
The ablation result is shown in Table~\ref{tab:hyper}.
As the ablation result demonstrates, the best choice of hyper-parameter $\lambda_{embed}$ should be $5$.

\begin{table}[t]
\renewcommand\arraystretch{1.5}
\caption{Ablation studies on CNN backbones.}
\label{tab:backbone}
\vspace{-10pt}
\begin{center}
\resizebox{1.0\linewidth}{0.08\textheight}{
\begin{tabular}{lcccc}
\midrule[1.0pt]
\rowcolor[gray]{0.8}Backbone & F1(\%)$\uparrow$ & ACC(\%)$\uparrow$ & FPR(\%)$\downarrow$ & FNR(\%)$\downarrow$\\ 
\midrule[1.0pt]
VGG16 & 90.83 & 90.27 & 7.94 & 8.15 \\
ResNet18 & 92.07 & 91.43 & 7.23 & 7.45\\
ResNet50 & 92.63 & \textbf{92.07} & 6.78 & 6.82 \\
ResNet101 & \textbf{92.91} & 91.95 & \textbf{6.45} & \textbf{6.58} \\
\midrule[1.0pt]
\end{tabular}}
\end{center}
\vspace{-10pt}
\end{table}

\noindent\textbf{Effects of backbones.}
To explore the effects of CNN backbones, we conduct the ablation study in Table~\ref{tab:backbone}.
The largest performance gap, with a difference of 2.08\% in F1 score, is observed between ResNet101~\cite{resnet} and VGG16~\cite{Simonyan2014VeryDC}.

\subsection{Visualization}
\noindent\textbf{Main visualizations.}
We have compared the quantified results of \textit{LaneCorrect} with supervised method on four major lane detection benchmarks.
In this section, we present visualization results on multiple datasets in Figure \ref{fig:main results}.

As is shown in Figure \ref{fig:main results}, our \textit{LaneCorrect} method demonstrates considerable performance compared with existing supervised counterpart.
The top and bottom row present the performance on \textit{TuSimple} and \textit{LLAMAS}.
The rest middle rows show qualitative results on \textit{CULane} in different scenarios.
The visulizations further prove that our \textit{LaneCorrect} method achieves considerable results in multiple benchmarks and various driving scenarios, including highway, urban, curves, night and others, which indicates the generalizability of the proposed approach.

\noindent\textbf{Effects of \textit{SLC} network.}
In the ablation study section of the main paper, we have explored the impact of the \textit{SLC} network.
We now present some qualitative results regarding the impact of the \textit{SLC} network.

As Figure~\ref{fig:nlc} demonstrates, with the help of self-supervised trained \textit{SLC} network, the performance of our unsupervised lane detector is substantially improved.
The visualization results are generated on \textit{Tusimple} da- \
taset.
The third column shows scenarios in which unsupervised \textit{na\"ive} lane detector predicts incorrectly, and the fourth column shows that with distillation of the self-supervised pre-trained \textit{SLC} network, our unsupervised lane detector is able to generate correct lane predictions in these cases where \textit{na\"ive} lane detector mispredicts.

The first row presents the missing prediction of our \textit{na\"ive} lane detector due to occlusion.
However, our \textit{SLC} network can detect all the lanes correctly.
The second row shows the case where \textit{na\"ive} lane detector incorrectly detects the curb as lane and the correct prediction of our method.
In the third row, we present our \textit{SLC} network can automatically refine the inaccurate lane locations predicted by unsupervised \textit{na\"ive} lane detector.

It is worth noting that, our \textit{SLC} network significantly improves the performance of the variant without \textit{SLC}, which proves the effectiveness of our self-supervised approach.

\noindent\textbf{Generalization.}
Here we show some qualitative results on cross domain task in Figure~\ref{fig:da}.
The third column presents the visualization results of supervised me- \
thod, PointLaneNet~\cite{pointlanenet}, which is strongly supervised trained on the \textit{CULane} training set and then directly test on \textit{TuSimple} test set.
To test the generalization performance of \textit{LULA}, we distill \textit{SLC} network to a student lane detector on \textit{CULane} and then directly evaluated on \textit{TuSimple} test set.
The performance is shown in the fourth column.

As is shown in Figure \ref{fig:da}, our \textit{LaneCorrect} method performs better than that of strongly supervised.
The qualitative results show that there are obvious cases of missing and false lane predictions, which proves the performance of supervised learning method is not ideal when transferred to other datasets.
One of the main reason is that the over-fitting of training datasets limits the generalization performance of the supervised model.
Our unsupervised method achieves excellent results on cross-domain tasks, which proves the generalization ability of our method.

\section{Conclusions}
We have shown the LiDAR view for the lane instance can be utilized as pseudo labels for lane detection task.
We have proposed a novel self-supervised training scheme for automatically correcting the noisy pseudo annotation with consistency and contrastive representation learning.
To perform lane detetion, we distill the pre-trained correction model to a student lane detector on arbitrary target lane benchmark without touching its ground-truth.
Our model achieves comparable performance with existing supervised counterpart (pre-trained on ImageNet~\cite{imagenet}) on four major benchmarks and shows superior performance on the domain adaptation setting.
We believe self-supervised learning method on lane detection can serve as a strong competitor to the supervised rivals for its scalability and generalizability.
The datasets generated during and/or analysed during the current study are available from the corresponding author on reasonable request.

\section*{Data Availability Statement}
The datasets generated during and/or analysed during the current study are available in the Waymo~\cite{sun2020scalability}, TuSimple~\cite{tusimple}, CULane~\cite{pan2018spatial}, LLAMAS~\cite{llamas2019} and CurveLanes~\cite{CurveLane-NAS}.

\section*{Acknowledgments}
This work was supported in part by National Natural Science Foundation of China (Grant No. 62376060).

\bibliographystyle{spbasic}      
\bibliography{main}   

\begin{thebibliography}{60}
\providecommand{\natexlab}[1]{#1}
\providecommand{\url}[1]{{#1}}
\providecommand{\urlprefix}{URL }
\expandafter\ifx\csname urlstyle\endcsname\relax
  \providecommand{\doi}[1]{DOI~\discretionary{}{}{}#1}\else
  \providecommand{\doi}{DOI~\discretionary{}{}{}\begingroup \urlstyle{rm}\Url}\fi
\providecommand{\eprint}[2][]{\url{#2}}

\bibitem[{Asano et~al.(2020)Asano, Rupprecht, and Vedaldi}]{Asano2020SelflabellingVS}
Asano YM, Rupprecht C, Vedaldi A (2020) Self-labelling via simultaneous clustering and representation learning. arXiv preprint

\bibitem[{Behrendt and Soussan(2019)}]{llamas2019}
Behrendt K, Soussan R (2019) Unsupervised labeled lane markers using maps. In: IEEE International Conference on Computer Vision

\bibitem[{Bhanu et~al.(1986)Bhanu, Lee, Ho, and Henderson}]{Bhanu1986RANGEDP}
Bhanu B, Lee S, Ho C, Henderson T (1986) Range data processing: Representation of surfaces by edges. In: IEEE International Conference on Pattern Recognition

\bibitem[{Cao et~al.(2019)Cao, Song, Xiao, and Peng}]{b-spline}
Cao J, Song S, Xiao W, Peng Z (2019) Lane detection algorithm for intelligent vehicles in complex road conditions and dynamic environments. Sensors

\bibitem[{Caron et~al.(2019)Caron, Bojanowski, Mairal, and Joulin}]{Caron2019LeveragingLU}
Caron M, Bojanowski P, Mairal J, Joulin A (2019) Leveraging large-scale uncurated data for unsupervised pre-training of visual features. CoRR

\bibitem[{Caron et~al.(2020)Caron, Misra, Mairal, Goyal, Bojanowski, and Joulin}]{Caron2020UnsupervisedLO}
Caron M, Misra I, Mairal J, Goyal P, Bojanowski P, Joulin A (2020) Unsupervised learning of visual features by contrasting cluster assignments. Advances in neural information processing systems

\bibitem[{Chen et~al.(2019{\natexlab{a}})Chen, Xie, Vedaldi, and Zisserman}]{Chen2019AutoCorrectDI}
Chen H, Xie W, Vedaldi A, Zisserman A (2019{\natexlab{a}}) Autocorrect: Deep inductive alignment of noisy geometric annotations. In: British Machine Vision Conference

\bibitem[{Chen et~al.(2020)Chen, Kornblith, Norouzi, and Hinton}]{Chen2020ASF}
Chen T, Kornblith S, Norouzi M, Hinton GE (2020) A simple framework for contrastive learning of visual representations. In: International Conference on Machine Learning

\bibitem[{Chen et~al.(2019{\natexlab{b}})Chen, Liu, and Lian}]{pointlanenet}
Chen Z, Liu Q, Lian C (2019{\natexlab{b}}) Pointlanenet: Efficient end-to-end cnns for accurate real-time lane detection

\bibitem[{Deng et~al.(2009)Deng, Dong, Socher, Li, Li, and Fei-Fei}]{imagenet}
Deng J, Dong W, Socher R, Li LJ, Li K, Fei-Fei L (2009) Imagenet: A large-scale hierarchical image database

\bibitem[{Dong et~al.(2018)Dong, Yang, Hu, and Scherer}]{DONG2018112}
Dong Z, Yang B, Hu P, Scherer S (2018) An efficient global energy optimization approach for robust 3d plane segmentation of point clouds. ISPRS Journal of Photogrammetry and Remote Sensing

\bibitem[{Ferraz et~al.(2010)Ferraz, Bretar, Jacquemoud, Gonçalves, and Pereira}]{5651310}
Ferraz A, Bretar F, Jacquemoud S, Gonçalves G, Pereira L (2010) 3d segmentation of forest structure using a mean-shift based algorithm. In: IEEE International Conference on Image Processing

\bibitem[{Fischler and Bolles(1981)}]{Fischler1981RandomSC}
Fischler M, Bolles R (1981) Random sample consensus: a paradigm for model fitting with applications to image analysis and automated cartography. Commun ACM

\bibitem[{Gao et~al.(2022)Gao, Ma, Li, Lin, Dai, and Qiao}]{gao2022mcmae}
Gao P, Ma T, Li H, Lin Z, Dai J, Qiao Y (2022) Mcmae: Masked convolution meets masked autoencoders. Advances in Neural Information Processing Systems

\bibitem[{Garnett et~al.(2020)Garnett, Uziel, Efrat, and Levi}]{garnett2020synthetic}
Garnett N, Uziel R, Efrat N, Levi D (2020) Synthetic-to-real domain adaptation for lane detection. In: accv

\bibitem[{Goyal et~al.(2021)Goyal, Caron, Lefaudeux, Xu, Wang, Pai, Singh, Liptchinsky, Misra, Joulin, and Bojanowski}]{Goyal2021SelfsupervisedPO}
Goyal P, Caron M, Lefaudeux B, Xu M, Wang P, Pai V, Singh M, Liptchinsky V, Misra I, Joulin A, Bojanowski P (2021) Self-supervised pretraining of visual features in the wild. arXiv preprint

\bibitem[{Hadsell et~al.(2006)Hadsell, Chopra, and LeCun}]{1640964}
Hadsell R, Chopra S, LeCun Y (2006) Dimensionality reduction by learning an invariant mapping. In: IEEE Conference on Computer Vision and Pattern Recognition

\bibitem[{He et~al.(2016)He, Zhang, Ren, and Sun}]{resnet}
He K, Zhang X, Ren S, Sun J (2016) Deep residual learning for image recognition. In: IEEE Conference on Computer Vision and Pattern Recognition

\bibitem[{He et~al.(2020)He, Fan, Wu, Xie, and Girshick}]{He2020MomentumCF}
He K, Fan H, Wu Y, Xie S, Girshick RB (2020) Momentum contrast for unsupervised visual representation learning. In: IEEE Conference on Computer Vision and Pattern Recognition

\bibitem[{He et~al.(2022)He, Chen, Xie, Li, Doll{\'a}r, and Girshick}]{he2022masked}
He K, Chen X, Xie S, Li Y, Doll{\'a}r P, Girshick R (2022) Masked autoencoders are scalable vision learners. In: CVPR

\bibitem[{H{\'e}naff et~al.(2021)H{\'e}naff, Koppula, Alayrac, Van~den Oord, Vinyals, and Carreira}]{Henaff2021EfficientVP}
H{\'e}naff OJ, Koppula S, Alayrac JB, Van~den Oord A, Vinyals O, Carreira J (2021) Efficient visual pretraining with contrastive detection. In: ICCV

\bibitem[{Hou et~al.(2019)Hou, Ma, Liu, and Loy}]{SAD}
Hou Y, Ma Z, Liu C, Loy CC (2019) Learning lightweight lane detection cnns by self attention distillation. In: IEEE International Conference on Computer Vision

\bibitem[{Hough(1962)}]{osti_4746348}
Hough PV (1962) Method and means for recognizing complex patterns. Uspatent

\bibitem[{Jiang et~al.(2018)Jiang, Zhou, Leung, Li, and Fei-Fei}]{Jiang2018MentorNetLD}
Jiang L, Zhou Z, Leung T, Li LJ, Fei-Fei L (2018) Mentornet: Learning data-driven curriculum for very deep neural networks on corrupted labels. In: International Conference on Machine Learning

\bibitem[{Jiang et~al.(1996)Jiang, Meier, and Bunke}]{572006}
Jiang X, Meier U, Bunke H (1996) Fast range image segmentation using high-level segmentation primitives. In: IEEE Winter Conference on Applications of Computer Vision

\bibitem[{Li et~al.(2019)Li, Li, Hu, and Yang}]{li2019line}
Li X, Li J, Hu X, Yang J (2019) Line-cnn: End-to-end traffic line detection with line proposal unit. TIP

\bibitem[{Lin et~al.(2021)Lin, Pintea, and van Gemert}]{lin2021semi}
Lin Y, Pintea SL, van Gemert J (2021) Semi-supervised lane detection with deep hough transform. In: ICIP

\bibitem[{Liu et~al.(2021)Liu, Chen, Zhu, and Tan}]{liu2021condlanenet}
Liu L, Chen X, Zhu S, Tan P (2021) Condlanenet: a top-to-down lane detection framework based on conditional convolution. In: ICCV

\bibitem[{Malach and Shalev-Shwartz(2017)}]{Malach2017DecouplingT}
Malach E, Shalev-Shwartz S (2017) Decoupling "when to update" from "how to update". In: Advances in Neural Information Processing Systems

\bibitem[{Mittal et~al.(2019)Mittal, Tatarchenko, and Brox}]{mittal2019s4}
Mittal S, Tatarchenko M, Brox T (2019) Semi-supervised semantic segmentation with high-and low-level consistency. IEEE transactions on pattern analysis and machine intelligence

\bibitem[{Nguyen and Le(2013)}]{inproceedings}
Nguyen A, Le B (2013) 3d point cloud segmentation: A survey

\bibitem[{Ning et~al.(2009)Ning, Zhang, Wang, and Jaeger}]{ning}
Ning X, Zhang X, Wang Y, Jaeger M (2009) Segmentation of architecture shape information from 3d point cloud

\bibitem[{Nishi et~al.(2021)Nishi, Ding, Rich, and H{\"o}llerer}]{Nishi2021AugmentationSF}
Nishi K, Ding Y, Rich A, H{\"o}llerer T (2021) Augmentation strategies for learning with noisy labels. arXiv preprint

\bibitem[{Ouali et~al.(2020)Ouali, Hudelot, and Tami}]{ouali2020overview}
Ouali Y, Hudelot C, Tami M (2020) An overview of deep semi-supervised learning. arXiv preprint

\bibitem[{Pan et~al.(2018{\natexlab{a}})Pan, Shi, Luo, Wang, and Tang}]{SCNN}
Pan X, Shi J, Luo P, Wang X, Tang X (2018{\natexlab{a}}) Spatial as deep: Spatial cnn for traffic scene understanding. In: AAAI Conference on Artificial Intelligence

\bibitem[{Pan et~al.(2018{\natexlab{b}})Pan, Shi, Luo, Wang, and Tang}]{pan2018spatial}
Pan X, Shi J, Luo P, Wang X, Tang X (2018{\natexlab{b}}) Spatial as deep: Spatial cnn for traffic scene understanding. In: AAAI Conference on Artificial Intelligence

\bibitem[{Patrini et~al.(2017)Patrini, Rozza, Menon, Nock, and Qu}]{Patrini2017MakingDN}
Patrini G, Rozza A, Menon A, Nock R, Qu L (2017) Making deep neural networks robust to label noise: A loss correction approach. In: IEEE Conference on Computer Vision and Pattern Recognition

\bibitem[{Qin et~al.(2020)Qin, Wang, and Li}]{qin2020ultra}
Qin Z, Wang H, Li X (2020) Ultra fast structure-aware deep lane detection. In: European Conference on Computer Vision

\bibitem[{Qu et~al.(2021)Qu, Jin, Zhou, Yang, and Zhang}]{Qu2021FocusOL}
Qu Z, Jin H, Zhou Y, Yang Z, Zhang W (2021) Focus on local: Detecting lane marker from bottom up via key point. arXiv preprint

\bibitem[{Rabbani et~al.(2006)Rabbani, Heuvel, and Vosselman}]{Rabbani}
Rabbani T, Heuvel F, Vosselman G (2006) Segmentation of point clouds using smoothness constraint. International Archives of Photogrammetry, Remote Sensing and Spatial Information Sciences

\bibitem[{Sampath and Shan(2010)}]{polyhedral}
Sampath A, Shan J (2010) Segmentation and reconstruction of polyhedral building roofs from aerial lidar point clouds. IEEE T Geoscience and Remote Sensing

\bibitem[{Shahzad et~al.(2012)Shahzad, Zhu, and Bamler}]{TomoSAR}
Shahzad M, Zhu X, Bamler R (2012) Façade structure reconstruction using spaceborne tomosar point clouds. International Geoscience and Remote Sensing Symposium

\bibitem[{Simonyan and Zisserman(2014)}]{Simonyan2014VeryDC}
Simonyan K, Zisserman A (2014) Very deep convolutional networks for large-scale image recognition. CoRR

\bibitem[{Sun et~al.(2020)Sun, Kretzschmar, Dotiwalla, Chouard, Patnaik, Tsui, Guo, Zhou, Chai, Caine et~al.}]{sun2020scalability}
Sun P, Kretzschmar H, Dotiwalla X, Chouard A, Patnaik V, Tsui P, Guo J, Zhou Y, Chai Y, Caine B, et~al. (2020) Scalability in perception for autonomous driving: Waymo open dataset. In: IEEE Conference on Computer Vision and Pattern Recognition

\bibitem[{Tabelini et~al.(2021{\natexlab{a}})Tabelini, Berriel, Paixao, Badue, De~Souza, and Oliveira-Santos}]{tabelini2021polylanenet}
Tabelini L, Berriel R, Paixao TM, Badue C, De~Souza AF, Oliveira-Santos T (2021{\natexlab{a}}) Polylanenet: Lane estimation via deep polynomial regression. In: IEEE International Conference on Pattern Recognition

\bibitem[{Tabelini et~al.(2021{\natexlab{b}})Tabelini, Berriel, Paix{\~a}o, Badue, Souza, and Oliveira-Santos}]{Tabelini2020KeepYE}
Tabelini L, Berriel R, Paix{\~a}o TM, Badue C, Souza AD, Oliveira-Santos T (2021{\natexlab{b}}) Keep your eyes on the lane: Real-time attention-guided lane detection. In: IEEE Conference on Computer Vision and Pattern Recognition

\bibitem[{Tian et~al.(2021)Tian, Chen, Dai, Zhang, and Zhu}]{tian2021unsupervised}
Tian H, Chen Y, Dai J, Zhang Z, Zhu X (2021) Unsupervised object detection with lidar clues. In: IEEE Conference on Computer Vision and Pattern Recognition

\bibitem[{TuSimple(2019)}]{tusimple}
TuSimple (2019) Tusimple benchmark. \url{https://github.com/TuSimple/tusimple-benchmark}

\bibitem[{Tóvári and Pfeifer(2012)}]{interpolation}
Tóvári D, Pfeifer N (2012) Segmentation based robust interpolation- a new approach to laser data filtering. International Archives of Photogrammetry, Remote Sensing and Spatial Information Sciences

\bibitem[{Van~Gansbeke et~al.(2021)Van~Gansbeke, Vandenhende, Georgoulis, and Van~Gool}]{Gansbeke2021UnsupervisedSS}
Van~Gansbeke W, Vandenhende S, Georgoulis S, Van~Gool L (2021) Unsupervised semantic segmentation by contrasting object mask proposals. In: ICCV

\bibitem[{Wu et~al.(2018)Wu, Xiong, Yu, and Lin}]{Wu2018UnsupervisedFL}
Wu Z, Xiong Y, Yu SX, Lin D (2018) Unsupervised feature learning via non-parametric instance discrimination. In: CVPR

\bibitem[{Xu et~al.(2020)Xu, Wang, Cai, Zhang, Liang, and Li}]{CurveLane-NAS}
Xu H, Wang SJ, Cai X, Zhang W, Liang X, Li Z (2020) Curvelane-nas: Unifying lane-sensitive architecture search and adaptive point blending. In: European Conference on Computer Vision

\bibitem[{Xu et~al.(2022)Xu, Cai, Zhao, Zhang, Xu, Fu, and Xue}]{xu2022rclane}
Xu S, Cai X, Zhao B, Zhang L, Xu H, Fu Y, Xue X (2022) Rclane: Relay chain prediction for lane detection. In: European Conference on Computer Vision, Springer, pp 461--477

\bibitem[{Xu et~al.(2018)Xu, Yao, Tuttas, Hoegner, and Stilla}]{8334807}
Xu Y, Yao W, Tuttas S, Hoegner L, Stilla U (2018) Unsupervised segmentation of point clouds from buildings using hierarchical clustering based on gestalt principles. IEEE Journal of Selected Topics in Applied Earth Observations and Remote Sensing

\bibitem[{Zeng et~al.(2019)Zeng, Zeng, and Kodom}]{hough-transform}
Zeng D, Zeng G, Kodom P (2019) Research on recognition technology of vehicle rolling line violation in highway based on visual uav. In: ICRCA

\bibitem[{Zhang and Maire(2020)}]{Zhang2020SelfSupervisedVR}
Zhang X, Maire M (2020) Self-supervised visual representation learning from hierarchical grouping. Advances in Neural Information Processing Systems

\bibitem[{Zhang et~al.(2021)Zhang, Zongqing, Zhang, Xue, and Liao}]{dllmd}
Zhang Y, Zongqing l, Zhang X, Xue JH, Liao Q (2021) Deep learning in lane marking detection: A survey. IEEE Transactions on Intelligent Transportation Systems

\bibitem[{Zheng et~al.(2022)Zheng, Huang, Liu, Tang, Yang, Cai, and He}]{zheng2022clrnet}
Zheng T, Huang Y, Liu Y, Tang W, Yang Z, Cai D, He X (2022) Clrnet: Cross layer refinement network for lane detection. In: CVPR

\bibitem[{Zhu and Shahzad(2014)}]{6573417}
Zhu XX, Shahzad M (2014) Facade reconstruction using multiview spaceborne tomosar point clouds. IEEE Transactions on Geoscience and Remote Sensing

\bibitem[{Zorzi et~al.(2020)Zorzi, Bittner, and Fraundorfer}]{map-repair}
Zorzi S, Bittner K, Fraundorfer F (2020) Map-repair: Deep cadastre maps alignment and temporal inconsistencies fix in satellite images. IEEE International Geoscience and Remote Sensing Symposium

\end{thebibliography}


\end{document}